\documentclass{article}
\usepackage{spconf,amsmath,graphicx} 
\usepackage{xcolor} % Add this package for colors
\usepackage{booktabs}
\usepackage{tikz}
\usetikzlibrary{patterns} 
\usepgflibrary{patterns}
\usepackage{pgfplots}
\pgfplotsset{compat=1.18} % Adjust compatibility version as needed
\usepackage{subcaption} % Add this to your preamble
\usepackage{hyperref}
\usepackage{caption}
\usepackage{graphicx}
\usepackage{geometry}
\usepackage{multirow}
\usepackage{enumitem}

\definecolor{visionblue}{rgb}{0.69, 0.83, 0.99} 
\definecolor{audiored}{rgb}{1.0, 0.45, 0.42} 
\definecolor{textgreen}{rgb}{0.37, 0.70, 0.16} 
\definecolor{audiovisionmix}{rgb}{0.43, 0.29, 0.39} 

\definecolor{audiotextmix}{rgb}{0.87, 0.75, 0.25}

\definecolor{darkgreen}{rgb}{0.0, 0.5, 0.0}

\title{
Leveraging multimodal explanatory annotations for video interpretation with Modality Specific Dataset}

\name{Elisa Ancarani$^{1}$, Julie Tores$^{1}$, Lucile Sassatelli$^{1}$, Rémy Sun$^{1,2}$, Hui-Yin Wu$^{2}$, Frédéric Precioso$^{1,2}$\thanks{This work was supported by the French National Research Agency ANR TRACTIVE project ANR-21-CE38-0012-01.}}
\address{$^{1}$Université Côte d’Azur, CNRS, I3S, France \hspace{.5cm} $^{2}$Université Côte d’Azur, Inria, France}

\begin{document}
%\ninept
%
\maketitle

\begin{abstract}

We examine the impact of concept-informed supervision on multimodal video interpretation models using MOByGaze, a dataset containing human-annotated explanatory concepts. We introduce Concept Modality Specific Datasets (CMSDs), which consist of data subsets categorized by the modality (visual, textual, or audio) of annotated concepts. Models trained on CMSDs outperform those using traditional legacy training in both early and late fusion approaches. Notably, this approach enables late fusion models to achieve performance close to that of early fusion models. These findings underscore the importance of modality-specific annotations in developing robust, self-explainable video models and contribute to advancing interpretable multimodal learning in complex video analysis.

\end{abstract}

\section{Introduction}\label{sec:intro}

In analysis of visual data, different types of tasks require different levels of human expertise. For example, object recognition or semantic segmentation are often less challenging for a human than assessing whether a skin lesion is malignant \cite{skincon} or a meme is offensive \cite{LaRochelle}.

For such complex image analysis tasks, which we call image interpretation, it is essential to have explainable models where the patterns behind decisions remain accessible and auditable to human experts.

Towards this goal, some image analysis datasets have been created with human experts providing a label for a complex interpretive task, along with the explanatory factors for their labeling decision. These are referred to as concepts. For example, CUB is an image dataset of 200 labeled bird species and annotated with 312 concepts attributes by experts \cite{cub}, while SkinCon is an dataset of skin lesion images labeled as malignant or benign along with 48 explanatory concepts. Such concept annotations as ground-truth explanations allow both to assess the explainability of models and design more efficient explainable models.

In video, datasets with concept annotations are rare owing to the tedious annotation process. Tores et al. recently introduced a dataset for a challenging video interpretation task: detecting whether a character in a movie clip is objectified—that is, portrayed as an object of desire or service rather than as an active subject \cite{cvpr}. A prominent aspect of this video dataset is the dense annotation of 20 movies by experts with concept-based explanations for their judgments.

\begin{figure}[tbp]
    \centering
    % First Figure
    \begin{minipage}[b]{\linewidth}
        \hfill(a)\hspace{0.5em}
        \includegraphics[width=0.9\linewidth]{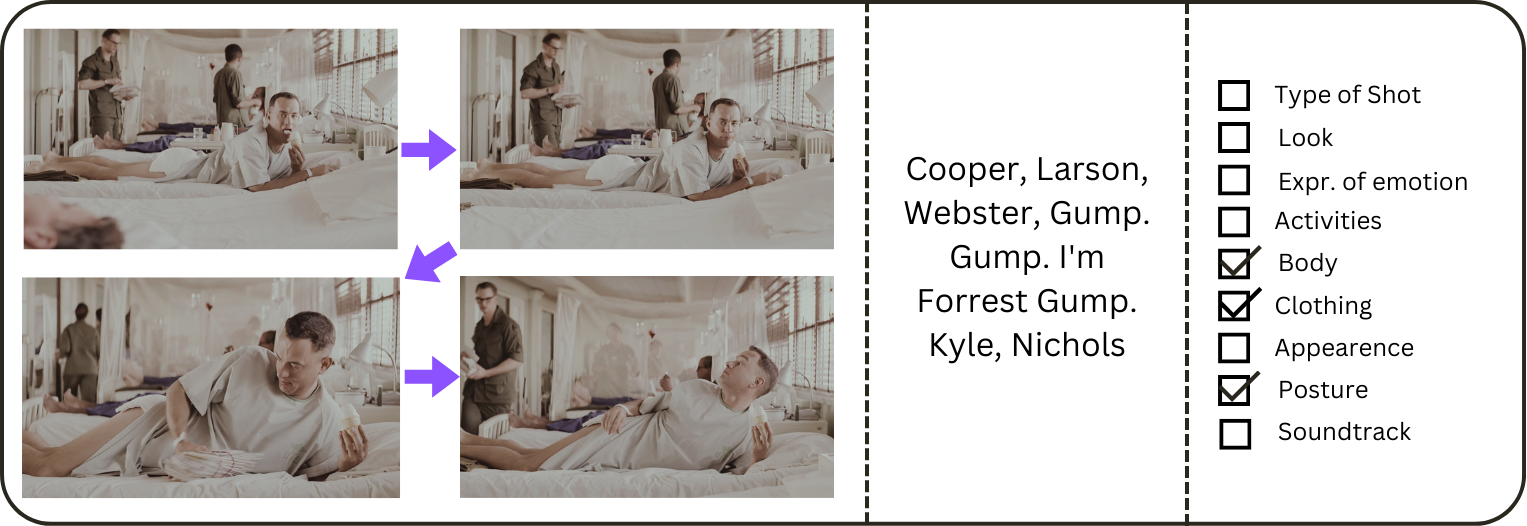}
    \end{minipage}
    \vspace{1em} % Add some vertical space between the figures
    % Second Figure
    \begin{minipage}[b]{\linewidth}
        \hfill(b)\hspace{0.5em}
        \includegraphics[width=0.9\linewidth]{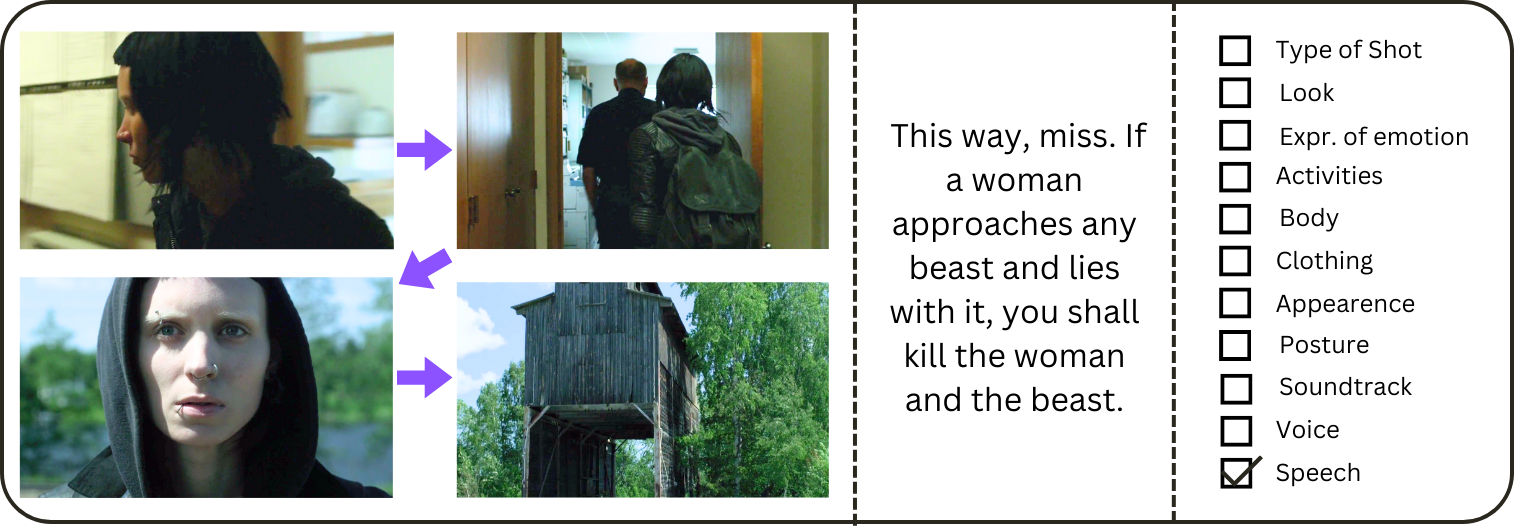}
    \end{minipage}
    \caption{
    Two positive samples for character objectification from the MObyGaze dataset. The concepts annotated indicate that objectification is due to the visual modality only in (a), and to the textual modality only in (b). For sample (a) (resp. (b)), the model fed with text only (resp. vision only) and trained with the Modality Agnostic Dataset wrongly detects objectification, while it correctly classifies the sample as negative when trained on the text-specific dataset T-CMSD (resp. vision-specific dataset V-CMSD).}

    \label{fig:example_combined}
\end{figure}

For such interpretive tasks in image or video, models must allow analysis of each modality’s influence on the final decision to be deemed trustworthy and fulfill their purpose, such as providing objective measures of bias in visual media.

In this article, we consider video interpretation and investigate whether and how annotation of explanations as concepts can benefit multimodal models, to eventually produce better explainable models. We specifically focus on the power of models to discriminate between what modalities explain the final label, as exemplified in Fig. \ref{fig:example_combined}. To do so, we consider two types of multimodal fusion, early and late fusion.

In video, several recent works address visual tasks informed by other modalities, like automatic auto-description \cite{autoad}, video-question answering \cite{llamavid} or social message detection \cite{mmaupaper}. These works build on early fusion of modalities, where tokens of different modalities are fed to a common Transformer network. This type of fusion however post-hoc explainability methods, such as saliency maps of the input data, whose brittleness has been recurrently shown \cite{rudin}. Late fusion on the other hand is required in self-explainable neural network models (e.g., \cite{SENN, whitening, proto, CBM, CEM}), where the production of the final decision (based on the importance of meaningful higher-level concepts) is directly interpretable by humans.

\textbf{Our contributions are}:\\
\noindent$\bullet$ We consider a recent video dataset with human annotations of multimodal (visual, textual, audio) explanatory concepts for a task of video interpretation. We design an assessment strategy with so-called Concept Modality Specific Datasets (CMSDs), which are sorted subsets based on annotated concept modalities. We assess the discriminative power of multimodal models when trained with such modality-informed supervision, in comparison with legacy training with Concept Modality Agnostic Dataset (CMAD).\\
\noindent$\bullet$ We show that training with CMSD significantly improves the performance of both late and early fusion models.\\
\noindent$\bullet$ We show that modality-informed supervision allow late fusion models to perform as well as early fusion, thereby showing the importance of modality-detailed annotation to enable high-performing self-explainable models for video interpretation tasks.

\section{Related works}\label{sec:related_works}

Our research focuses on video interpretation at the intersection of two areas: (1) concept-annotated datasets and (2) multimodal fusion. As we show below, we are the first to explore concept annotations in multimodal training to enhance model performance.

\noindent\textbf{Concept-annotated Datasets} Many image and video datasets are anotated (often partially automatically) with multiple semantic aspects. For example, 3MASSIV is a multilingual, multimodal dataset of short social media videos annotated with concepts, affective states, media types, and audio language. However in this article, we consider concept annotations as ground-truth explanations provided by humans on their labeling decision for an end interpretive task. Example of such image datasets are CUB \cite{cub} and SkinCon \cite{skincon} introduced in Sec. \ref{sec:intro}. 
Video interpretation tasks are rarer. A recent dataset is MM-AU \cite{mmaupaper}, of 8399 advertisement videos annotated by humans for three interpretive tasks: topic categorization, perceived tone transition, and social message detection. However, MM-AU does not provide annotations specifying whether the end label depends on some or all modalities (visual, textual, or audio). More recently, Tores et al. introduced the ObyGaze dataset \cite{cvpr}, where movie clips have been densely annotated by experts for the presence of character objectification. As this is a high-level interpretive construct, annotators were required to provide explanations for their judgment. For each label of objectification presence or absence, they indicate wheteher it is due to one or several of the 8 visual concepts (such as body, clothing, type of shot, etc.).
In our work, we consider the multimodal extension of this dataset, called MObyGaze, which further described in Sec. \ref{sec:method}. It features 20 films annotated with 11 concepts spanning all three modalities (visual, textual and audio). We therefore explore how to best learn from this unique dataset tying multimodal explanations to an interpretive task label. \\
\noindent\textbf{Multimodal Fusion} 
Fusing information from multiple modalities is a frequent challenge in numerous tasks (such as violence detection in multimedia content \cite{violencedetection} or intent recognition \cite{intent_recognition}). We can roughly classify the approaches into early and late fusion. Early fusion combines raw data from different modalities at the input level, while late fusion processes each modality independently before combining outputs. Early fusion has demonstrated increased robustness to noisy inputs, as information from one modality can help mitigate noise in another \cite{onbeefitsofearlyfsuion}. It also effectively exploits cross-modal correlations when the data is well-aligned \cite{earlyvslatefusion}.
Early fusion is however not amenable to explainability approaches beyond post-hoc methods.
late fusion, on the other hand, preserves distinct modalities until or close to the final decision (in final layers). This allows more reliable interpretations of each modality’s contribution to the final outcome (see, e.g., SENN \cite{SENN}), but comes at the cost of lower resilience to noise—particularly when a modality is less informative or introduces interference \cite{onbeefitsofearlyfsuion}.

When introducing the above-mentioned MM-AU dataset, Gupta et al. explore both unimodal and multimodal baselines, combining different modality pairs using a mix of early and late fusion. Specifically, they apply early fusion to the vision-text and vision-audio modality pairs, while performing late fusion by combining class probabilities at inference. Given the similarity of their tasks to objectification tasks and the availability of concept annotations, we build on their models to investigate early fusion, focusing exclusively on early fusion of modality pairs.

\section{Method}\label{sec:method}

We present here our method to investigate how concept annotations can benefit multimodal architectures for video interpretation, and whether early and late fusion benefit differently.\\

\noindent\textbf{Case study}: As introduced in Sec. \ref{sec:intro}, we consider detecting objectification in videos, which occurs when a character is portrayed as an object of desire or service rather than a subject of action (often referred to as the ``male gaze''). This is a task recently introduced by Tores et al. \cite{cvpr} with the ObyGaze12 dataset.
Here we consider its multimodal extension, the MObyGaze dataset available online\footnote{\href{https://anonymous.4open.science/r/MObyGaze-F600/}{https://anonymous.4open.science/r/MObyGaze-F600/}}.
Objectification is produced by a combination of cinematic techniques (camera position, angle, movement), iconographic choices (visible body parts, clothing, looks, etc.), narrative (speech) and auditory components (voice and soundtrack). 
MObyGaze comprises 5,783 video clips from 20 movies densely annotated by human experts with objectification levels: Easy Negative (EN) where no objectifying factor is present, Hard Negative (HN) where at least one objectifying factor/concept is present, but not sufficient to fully perceive the character as an object, and Sure (S) where objectifying concepts are sufficiently present to produce the perception of objectification. 
In this article, as we focus on identifying modalities involved in the production of objectification, we consider HN and S samples as positive, as further described below.%in 

\begin{figure}[ht!]
    \centering
    \includegraphics[width=0.35\textwidth, height=0.25\textwidth]{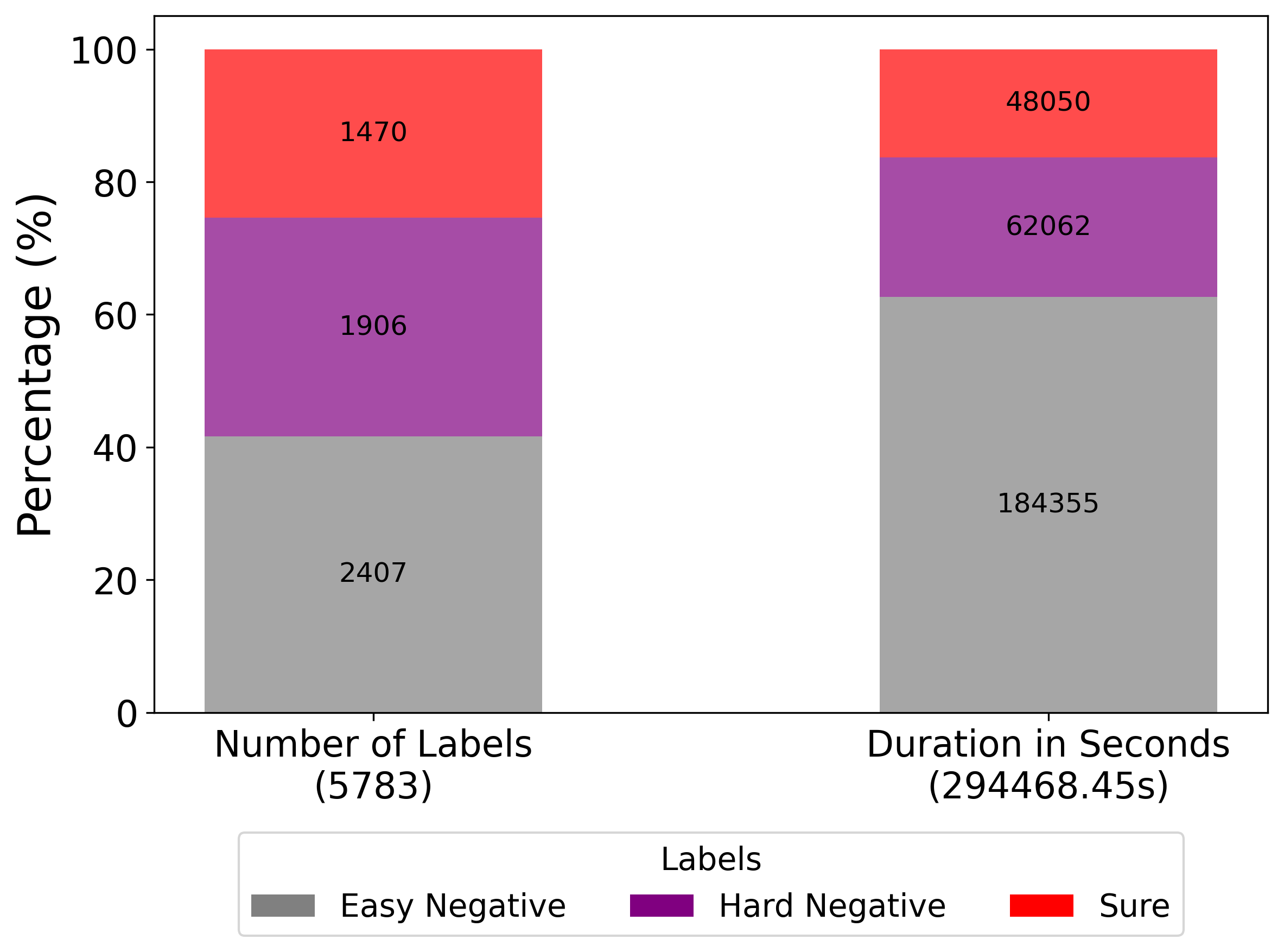} 
    \caption{Distribution of objectifying labels by frequency and duration in the MObyGaze dataset.}\label{fig:number_of_label_2}
\end{figure}

\begin{figure}[ht!]
    \centering
    \includegraphics[width=0.45\textwidth, height=0.30\textwidth]{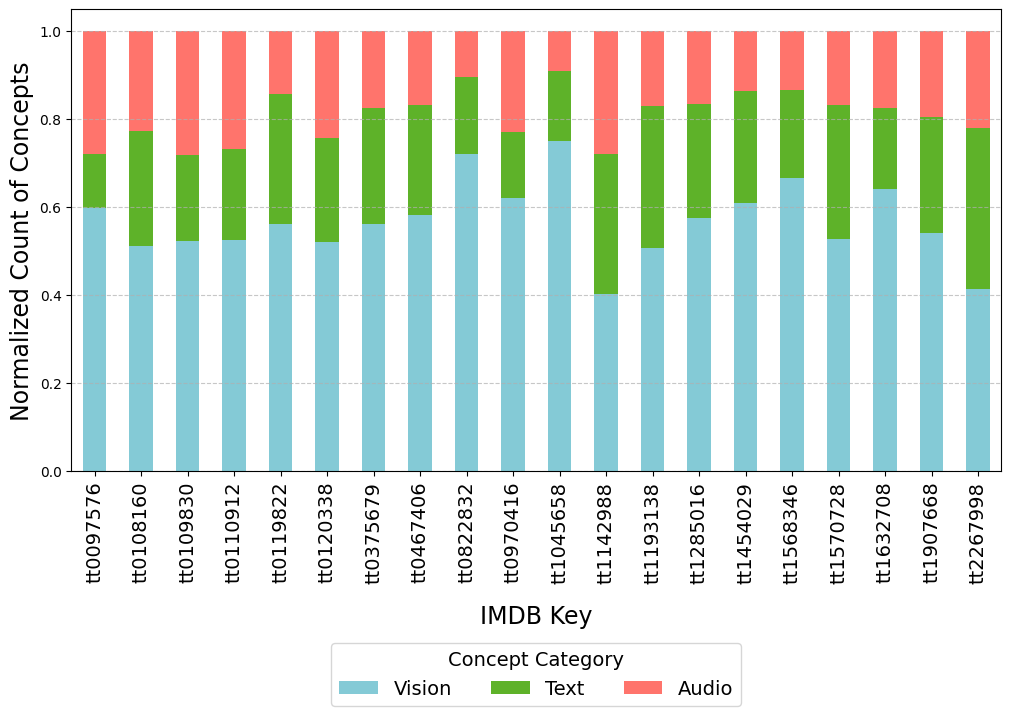}
    \caption{Normalized distribution of visual, text, and audio concepts across movies.}
    \label{fig:modality_distribution}
    \vspace{-0.5cm}
\end{figure}

Fig. \ref{fig:number_of_label_2} illustrates the distribution of objectification labels across movies. 
Fig. \ref{fig:modality_distribution} showcases the distribution of concept modalities for each movie. Vision emerges as the dominant modality. This highlights this dataset's particular suitability for visual interpretation tasks.

\noindent\textbf{Assessment Strategy}: We focus on modality attribution errors, where objectification is attributed to an incorrect source compared to the ground truth annotations. 
A clip is labeled as positive (HN or S) if it contains at least one objectifying concept in any of the visual, audio, or textual modalities. For example, as shown in Fig. \ref{fig:example_combined}, in case (a), a model fed with text only may wrongly classify a sample as objectifying, even though concept annotations indicate that objectification is only produced by the visual modality. Such discrepancies reveal the model exploitation of undesired biases in the data, compromising trustworthiness and explainability. To probe model weaknesses and leverage concept annotations, we define two types of datasets from MObyGaze:

$\bullet$ \noindent\textbf{Concept Modality Agnostic Dataset (CMAD)}: This is the standard MOByGaze split with all positive clips (HN and S) based on speech, text, and/or audio, regardless of input modality, while negative clips consist only of EN clips. This follows the CMAD approach, which considers all objectifying samples without distinguishing the modality of objectification.

\begin{figure}[ht!]
    \centering
    \includegraphics[scale=0.2]{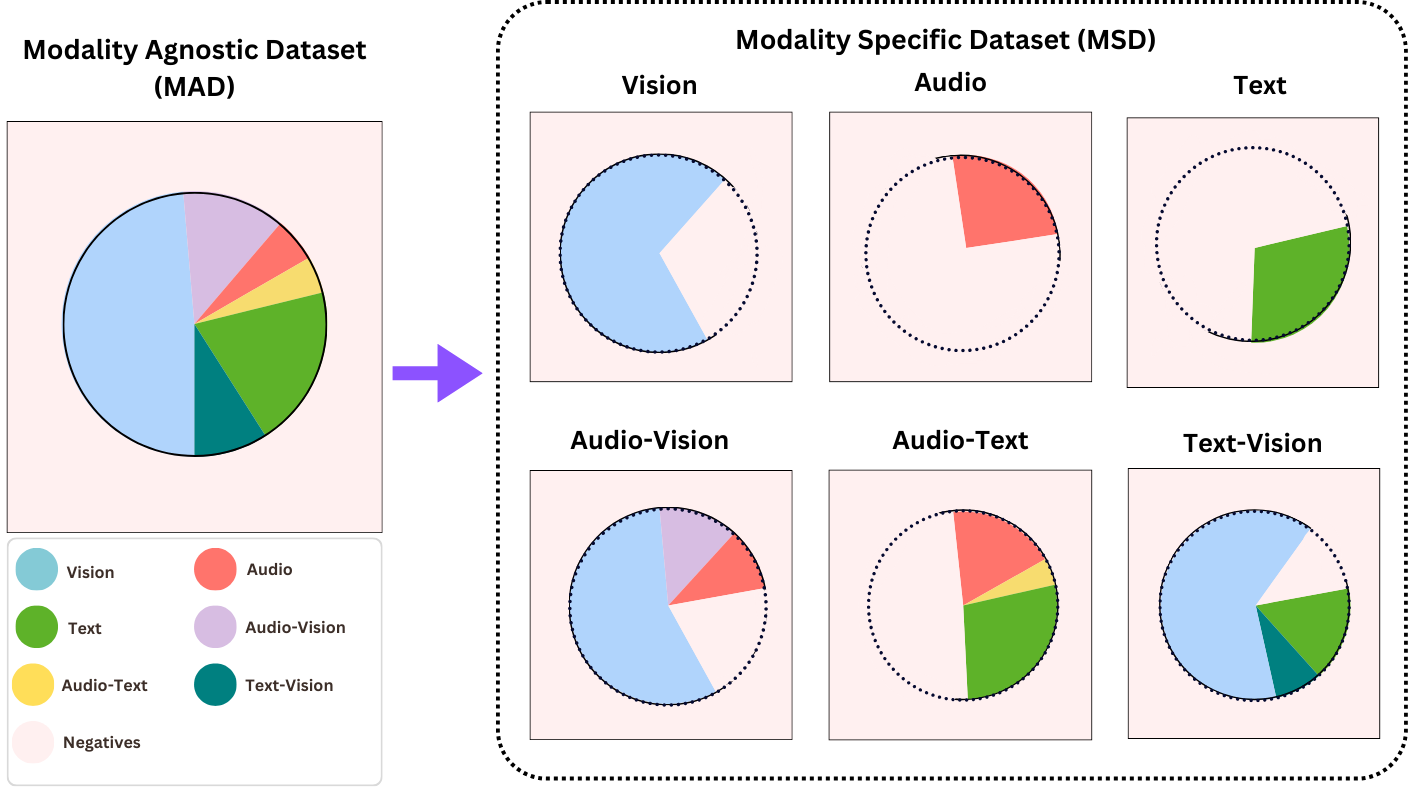} % Adjusted size
    \caption{The Concept Modality Agnostic Dataset (CMAD) contains all positive samples (Hard Negative and Sure) across modalities, while the Concept Modality Specific Dataset (CMSD) filters these samples to include only those with modality-specific concepts.}
    \label{fig:pipeline_data}
\end{figure}

$\bullet$ \noindent\textbf{Concept Modality Specific Dataset (CMSD)}: We propose CMSD as a concept-based supervision dataset. The positive set includes HN and S clips with at least one annotated concept of the model input modality. The negative set consists of EN clips together with the clips that do not meet the positive criteria. As shown in Fig. \ref{fig:pipeline_data}, CMSD is derived from CMAD by filtering positive samples to retain only modality-relevant concepts. 

As outlined in Table \ref{tab:models_table}, all models are evaluated on their respective CMSD test set. However, they can be trained separately using either CMAD or the CMSD specific to their input modality. For example, Vision-only models are evaluated on Vision-CMSD, which contains as positives the HN and S of the CMAD that contains at least one concept related to the vision modality. These models can however be trained either on CMAD or on Vision-CMSD. This strategy allows for a comprehensive assessment of the models' ability to disambiguate the source of objectification across modalities, depending on whether supervision leverages concept information.

We exclude Text-Vision pairing from our analysis as it represents almost the entire dataset. Therefore, it would limit our ability to effectively discriminate between results of CMAD and CMSD models.

\begin{table}[ht]
\centering
\footnotesize
\begin{tabular}{l|c|c}
\toprule
\textbf{Input Modalities} & \textbf{Training Sets} & \textbf{Test Sets} \\
\midrule
Vision              & CMAD or \textcolor{visionblue}{V-CMSD}  & \textcolor{visionblue}{V-CMSD} \\
Audio              & CMAD or \textcolor{audiored}{A-CMSD}   & \textcolor{audiored}{A-CMSD} \\
Text               & CMAD or \textcolor{textgreen}{T-CMSD}  & \textcolor{textgreen}{T-CMSD} \\
Audio-Vision      & CMAD or \textcolor{audiovisionmix}{AV-CMSD} & \textcolor{audiovisionmix}{AV-CMSD} \\
Audio-Text       & CMAD or \textcolor{audiotextmix}{AT-CMSD}  & \textcolor{audiotextmix}{AT-CMSD} \\
Vision-Audio-Text & CMAD or AVT-CMSD  & AVT-CMSD \\
\bottomrule
\end{tabular}
\caption{Training and testing sets for models with different input modalities.}
\label{tab:models_table}
\end{table}

\noindent\textbf{Models}: To investigate how different model designs can benefit from CMSD training (thanks to concept information) compared with legacy CMAD training, we consider early and late fusion of modalities. 
First, we consider \textbf{Unimodal Models} fed with a single modality, processing either vision, audio, or text. Each such model is a set of Transformer layers fed with modality tokens obtained from pre-trained models (see Sec. \ref{sec:experiments}). Unimodal models allow to assess the benefit of the CMSD strategy regardless of any mulitmodal fusion choice. 

\begin{figure}[ht!]
    \centering
    \begin{subfigure}[b]{0.45\textwidth}
        \centering
        \includegraphics[height=2.5cm]{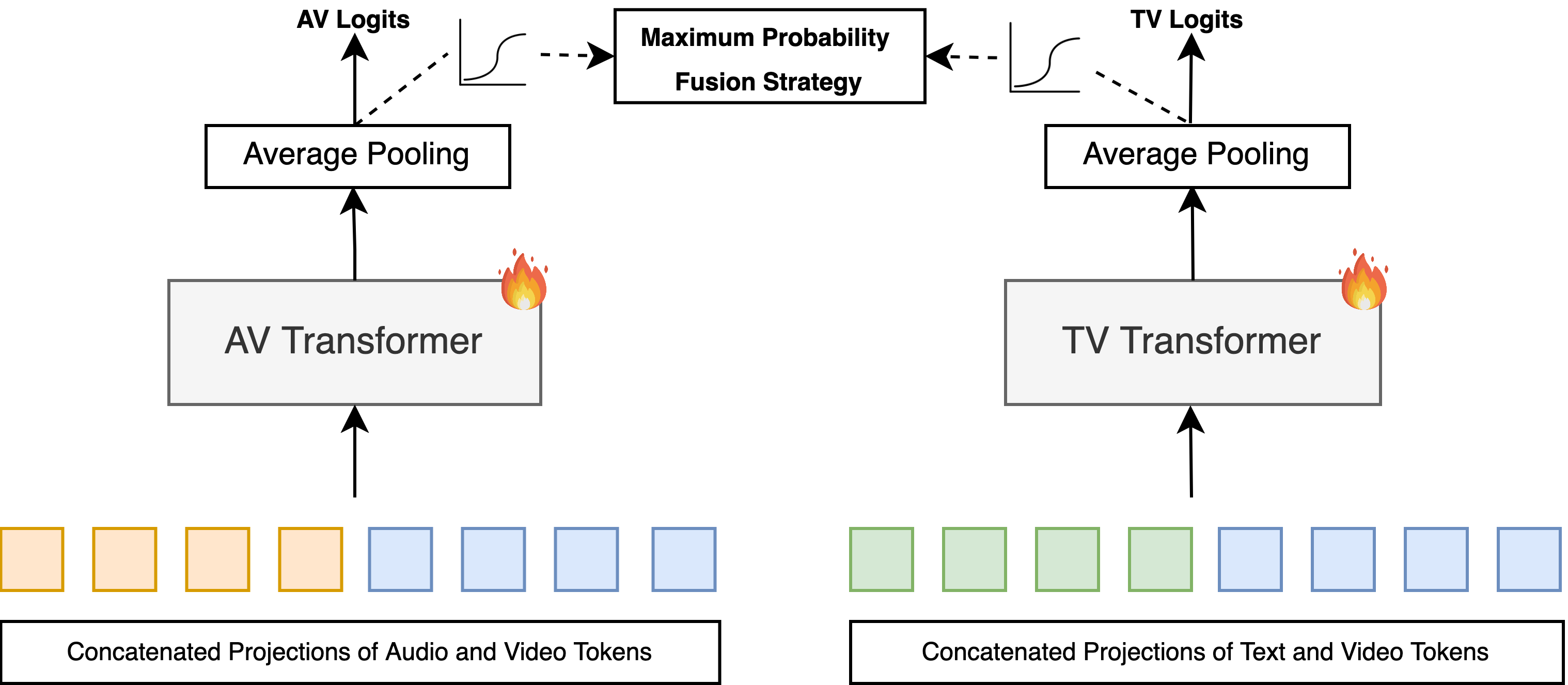} % Reduced height
        \caption{Early Fusion Model}
        \label{fig:early_fusion_model}
    \end{subfigure}
    
    \vspace{0.5cm} 

    \begin{subfigure}[b]{0.45\textwidth}
        \centering
        \includegraphics[height=3cm]{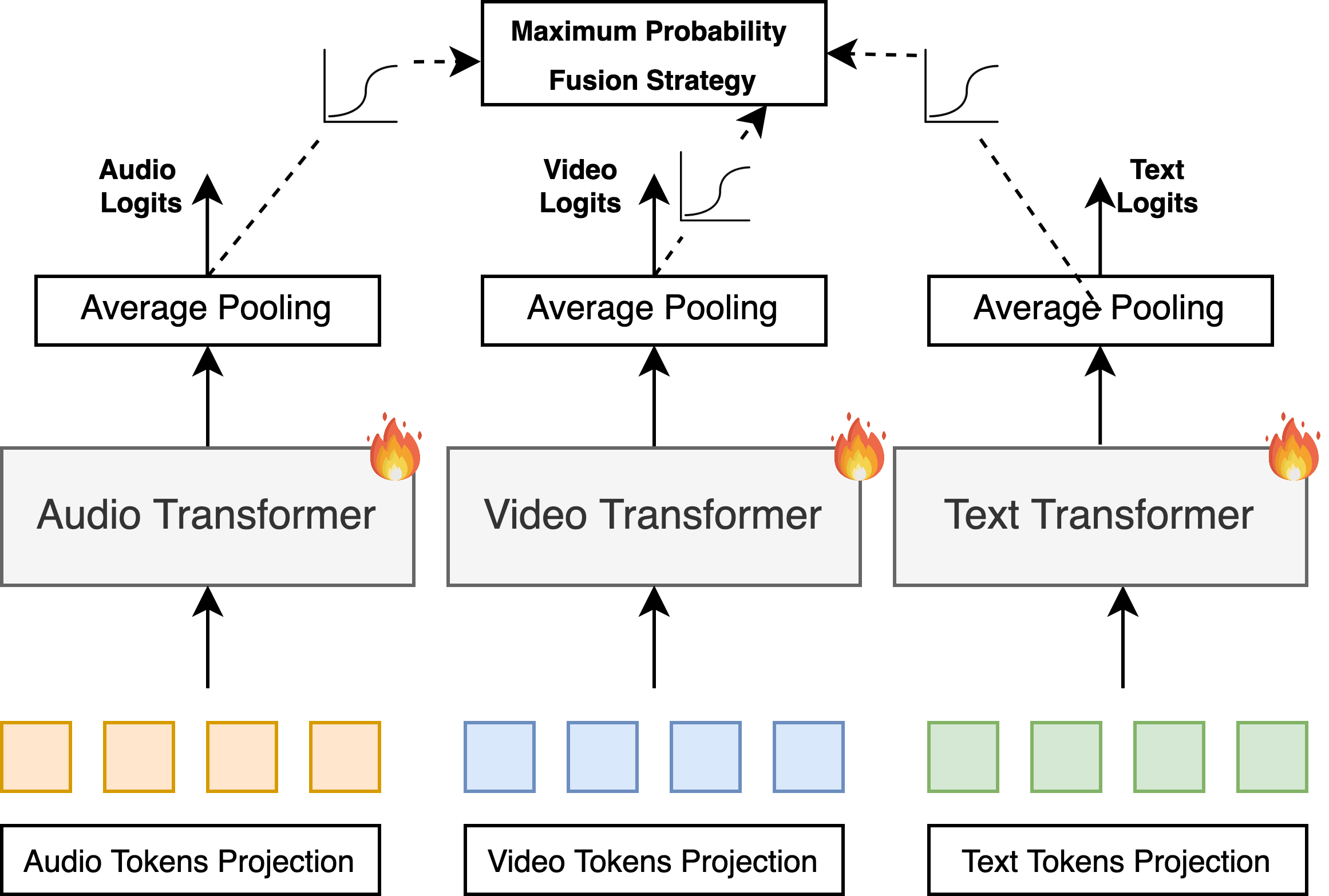} % Reduced height
        \caption{Late Fusion Model}
        \label{fig:late_fusion_model}
    \end{subfigure}
    \caption{\textbf{Comparison of Early and Late Fusion Models considering all three modalities}. Tokens are: Audio (orange), Video (blue), and Text (green).
    }
    \label{fig:fusion_models}
    \vspace{-0.5cm}
\end{figure}

Second, we consider \textbf{Mulitmodal Models} to study the interaction between fine-grained supervision and fusion. We focus on two main strategies: early and late fusion. 
Our choice for \textbf{early fusion} is shown in Fig. \ref{fig:early_fusion_model}. We build on the transformer-based framework from \cite{mmaupaper}. Token projections of paired modalities are processed jointly within two shared transformer models: one combines vision and text tokens, the other vision and audio tokens. The two models are trained separately. For the early fusion model with three input modalities, the output probabilities of each pair model are then max-fused at inference, as depicted.

For late fusion, we consider three modality-specific models, each fed with a single modality as shown in Fig. \ref{fig:late_fusion_model}. They are trained independently, and their respective output probabilities are max-fused at inference time, being it for the case of the late fusion model with three input modalities depicted in Fig. \ref{fig:late_fusion_model}, or when only two modalities are considered.

\section{Experiments}\label{sec:experiments}

We investigate experimentally ((Research Question - RQ1) whether training on our CMSD dataset splits enhances model performances, and (RQ2) we look into whether late and early fusion schemes benefit differently from CMSD training.

\noindent\textbf{Dataset}: 
We perform experiments on the MObyGaze dataset discussed in previous sections.

\noindent\textbf{Model settings}: To preprocess the visual modality, we leverage X-CLIP \cite{xclip}, an extension of CLIP \cite{clip} designed for videos. The pre-trained X-CLIP model is kept frozen, and we extract a feature vector from each 16-frame window with a stride of 16, for each annotated video segment. For textual modality, we extract textual cues from subtitles and encode them using the pretrained BERT model (bert-base-uncased) available in the Hugging Face library \cite{bert, huggingfacetransformers}.

For audio, we segment movie audio tracks into two-second chunks, and process them using the AST feature extractor \cite{ast}, pre-trained on AudioSet \cite{audioset}. For training, validation, and testing, we use a five-fold split across 20 movies. The training is done with random oversampling on the minority class. We use a validation set to stop the training with early stopping with patience of 15. 
The backbone architecture for all the unimodal and multimodal models is based on PerceiverIO \cite{PerceiverIO}, chosen for its ability to scale efficiently while handling high-dimensional inputs from multiple modalities. The model is configured with a depth of 2 layers, 32 latent variables, and a latent dimension of 128. It employs 1 cross-attention head and 8 latent self-attention heads, each with a dimensionality of 32. All models, have a comparable number of trainable parameters. In the early fusion method, each of the two model contains approximately 1 million parameters. Similarly, the three independent models in the late fusion also have about 1 million parameters each.

\noindent\textbf{Results} :  
To evaluate performance while accounting for class imbalance, we report AUC-PR scores. Our analysis focuses on three key dimensions: \vspace{-5pt}
\begin{itemize}[noitemsep]
    \setlength{\itemsep}{0pt}
    \setlength{\topsep}{0pt}
    \setlength{\parsep}{0pt}
    \item \textbf{Training}: Concept Modality-Agnostic Dataset (CMAD) vs. Modality-Specific Dataset (CMSD).
    \item \textbf{Modalities}: Vision, Audio, Text, Audio-Vision, Audio-Text, Vision-Audio-Text.
    \item \textbf{Types of fusion}: Early vs. Late fusion.
\end{itemize}  \vspace{-5pt}
Fig. \ref{fig:indiv_modalities}-\ref{fig:three_modalities} illustrate these dimensions, with each figure representing different modality sets. Color intensity indicates the training strategy (darker for CMSD, lighter for CMAD). Error bars indicate confidence intervals. 

Fig. \ref{fig:indiv_modalities} represents the performance of a Vision-only model (left) and Text-only model (right), when trained without CMAD or with CMSD (concept-based supervision). For both single modalities, we observe significant improvement with concept-based supervision: a $6.2\%$ increase in AUC-PR for Vision-only and a $63\%$ increase for Text-only. The substantial improvement in the Text-only model stems from the rarer occurrence of samples annotated with textual concepts compared to visual ones. This limited representation makes discriminating positive samples more challenging, yielding greater benefits from concept-informed supervision. Let us specify that, while we report AUC-PR for conciseness, we confirm that the increase in AUC-PR is mostly due to the increase in precision. This confirms the interest of concept-based supervision to mitigate the modality-attribution error exemplified in Fig. \ref{fig:example_combined}. \\
Fig. \ref{fig:paired_modalities} shows the performance of models using paired modalities as input: Vision-Audio (left) and Text-Audio (right). We compare early and late fusion approaches for each pair. Concept-based supervision significantly improves performance for both modality pairs and fusion types. Text-Audio models benefit more from concept-based supervision, with a $30.0\%$ increase in early fusion and $25.6\%$ increase in late fusion. In contrast, Vision-Audio models demonstrate more modest improvements, with increases of $3.72\%$ in early fusion and $3.92\%$ in late fusion. This pattern mirrors the observations in Fig. \ref{fig:indiv_modalities}. \\ 
Fig \ref{fig:three_modalities} reveals the impact of concept-based supervision on Early and Late fusion models combining all three modalities (vision, text, audio). In this scenario, CMSD equals CMAD, creating a less challenging discrimination task as the test set contains all samples from all modalities. 

Two key observations emerge from this analysis. First, early fusion demonstrates greater resilience to the absence of concept-based supervision compared to late fusion. This stems from early fusion's ability to learn multimodal features, which enhances its robustness across modalities. Second, concept-based supervision significantly narrows the performance gap between late fusion and early fusion models. Without concept-based supervision, late fusion performs $7.25\%$ worse than early fusion. However, with concept-based supervision, this gap reduces to just $3.30\%$, with overlapping confidence intervals. \\
These findings answer our initial research questions. RQ1: Concept-based supervision improves performance by reducing attribution errors across all models and modality sets. RQ2: With all three modalities combined, late fusion benefits most from concept-based supervision, closing the performance gap with early fusion.

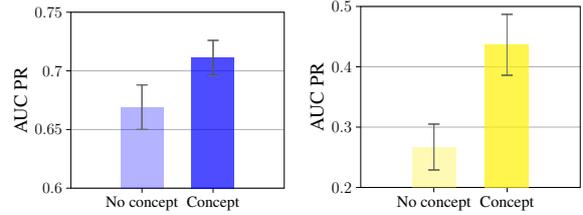
\begin{figure}[!ht]
    \centering
    % This file was created with tikzplotlib v0.10.1.
\resizebox{0.48\columnwidth}{!}{
\begin{tikzpicture}

\definecolor{darkgray176}{RGB}{176,176,176}
\definecolor{gray}{RGB}{128,128,128}
\definecolor{mediumseagreen7617580}{RGB}{76,175,80}
\definecolor{peachpuff}{RGB}{255,218,185}

\begin{axis}[
tick align=outside,
tick pos=left,
label style={font=\LARGE},
tick label style={font=\Large},
x grid style={darkgray176},
xmin=0.25, xmax=0.4,
xtick style={color=black},
xtick={0.3,0.35},
xticklabels={No concept,Concept},
y grid style={darkgray176},
ylabel={AUC PR},
ymajorgrids,
ymin=0.6, ymax=0.75,
ytick style={color=black}
]
\draw[draw=none,fill=blue,fill opacity=0.3] (axis cs:0.285,0) rectangle (axis cs:0.315,0.669);
%\addlegendimage{ybar,ybar legend,draw=none,fill=peachpuff,fill opacity=0.7}
%\addlegendentry{No Concept}

\draw[draw=none,fill=blue,fill opacity=0.7] (axis cs:0.335,0) rectangle (axis cs:0.365,0.711);
%\addlegendimage{ybar,ybar legend,draw=none,fill=mediumseagreen7617580,fill opacity=0.7}
%\addlegendentry{Concept}

\path [draw=black, semithick]
(axis cs:0.3,0.65)
--(axis cs:0.3,0.688);

\path [draw=black, semithick]
(axis cs:0.35,0.696)
--(axis cs:0.35,0.726);

\path [draw=gray, semithick]
(axis cs:0.3,0.65)
--(axis cs:0.3,0.688);

\path [draw=gray, semithick]
(axis cs:0.35,0.696)
--(axis cs:0.35,0.726);

\addplot [semithick, black, mark=-, mark size=5, mark options={solid}, only marks]
table[row sep=crcr]{%
0.3 0.65\\
};
\addplot [semithick, black, mark=-, mark size=5, mark options={solid}, only marks]
table[row sep=crcr]{%
0.3 0.688\\
};
\addplot [semithick, black, mark=-, mark size=5, mark options={solid}, only marks]
table[row sep=crcr]{%
0.35 0.696\\ 
};
\addplot [semithick, black, mark=-, mark size=5, mark options={solid}, only marks]
table[row sep=crcr]{%
0.35 0.726
\\ };
\addplot [semithick, gray, mark=-, mark size=5, mark options={solid}, only marks]
table[row sep=crcr]{%
0.3 0.65\\ 
};
\addplot [semithick, gray, mark=-, mark size=5, mark options={solid}, only marks]
table[row sep=crcr]{%
0.3 0.688\\ 
};
\addplot [semithick, gray, mark=-, mark size=5, mark options={solid}, only marks]
table[row sep=crcr]{%
0.35 0.696\\ 
};
\addplot [semithick, gray, mark=-, mark size=5, mark options={solid}, only marks]
table[row sep=crcr]{%
0.35 0.726\\ 
};

\end{axis}

\end{tikzpicture}
}
    \hfill
   % This file was created with tikzplotlib v0.10.1.
\resizebox{0.48\columnwidth}{!}{
\begin{tikzpicture}

\definecolor{darkgray176}{RGB}{176,176,176}
\definecolor{gray}{RGB}{128,128,128}
\definecolor{mediumseagreen7617580}{RGB}{76,175,80}
\definecolor{peachpuff}{RGB}{255,218,185}

\begin{axis}[
tick align=outside,
tick pos=left,
label style={font=\LARGE},
tick label style={font=\Large},
x grid style={darkgray176},
xmin=0.25, xmax=0.4,
xtick style={color=black},
xtick={0.3,0.35},
xticklabels={No concept,Concept},
y grid style={darkgray176},
ylabel={AUC PR},
ymajorgrids,
ymin=0.2, ymax=0.5,
ytick style={color=black}
]
\draw[draw=none,fill=yellow,fill opacity=0.3] (axis cs:0.285,0) rectangle (axis cs:0.315,0.267);
%\addlegendimage{ybar,ybar legend,draw=none,fill=peachpuff,fill opacity=0.7}
%\addlegendentry{No Concept}

\draw[draw=none,fill=yellow,fill opacity=0.7] (axis cs:0.335,0) rectangle (axis cs:0.365,0.4365);
%\addlegendimage{ybar,ybar legend,draw=none,fill=mediumseagreen7617580,fill opacity=0.7}
%\addlegendentry{Concept}

\path [draw=black, semithick]
(axis cs:0.3,0.229)
--(axis cs:0.3,0.305);

\path [draw=black, semithick]
(axis cs:0.35,0.386)
--(axis cs:0.35,0.487);

\path [draw=gray, semithick]
(axis cs:0.3,0.229)
--(axis cs:0.3,0.305);

\path [draw=gray, semithick]
(axis cs:0.35,0.386)
--(axis cs:0.35,0.487);

\addplot [semithick, black, mark=-, mark size=5, mark options={solid}, only marks]
table[row sep=crcr]{%
0.3 0.229\\
};
\addplot [semithick, black, mark=-, mark size=5, mark options={solid}, only marks]
table[row sep=crcr]{%
0.3 0.305\\
};
\addplot [semithick, black, mark=-, mark size=5, mark options={solid}, only marks]
table[row sep=crcr]{%
0.35 0.386\\
};
\addplot [semithick, black, mark=-, mark size=5, mark options={solid}, only marks]
table[row sep=crcr]{%
0.35 0.487\\
};
\addplot [semithick, gray, mark=-, mark size=5, mark options={solid}, only marks]
table[row sep=crcr]{%
0.3 0.229\\
};
\addplot [semithick, gray, mark=-, mark size=5, mark options={solid}, only marks]
table[row sep=crcr]{%
0.3 0.305\\
};
\addplot [semithick, gray, mark=-, mark size=5, mark options={solid}, only marks]
table[row sep=crcr]{%
0.35 0.386\\
};
\addplot [semithick, gray, mark=-, mark size=5, mark options={solid}, only marks]
table[row sep=crcr]{%
0.35 0.487\\
};

\end{axis}

\end{tikzpicture}
}
    \caption{Left: Video as single input modality. Right: Text as single input modality. } %\elisa{maybe it is better to have for both left and right bars the same colors}
    \label{fig:indiv_modalities}
    \vspace{-0.5cm}
\end{figure}
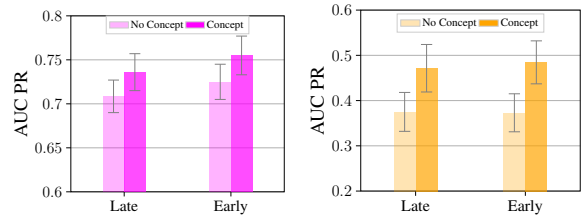
\begin{figure}[!ht]
    \centering
   % This file was created with tikzplotlib v0.10.1.
\resizebox{0.48\columnwidth}{!}{
\begin{tikzpicture}

\definecolor{darkgray176}{RGB}{176,176,176}
\definecolor{gray}{RGB}{128,128,128}
\definecolor{lightgray204}{RGB}{204,204,204}
\definecolor{magenta}{RGB}{255,0,255}

\begin{axis}[
legend cell align={left},
legend columns=2,
legend style={
  fill opacity=0.8,
  draw opacity=1,
  text opacity=1,
  at={(0.5,1)},
  anchor=north,
  draw=lightgray204
},
tick align=outside,
tick pos=left,
label style={font=\LARGE},
tick label style={font=\Large},
x grid style={darkgray176},
xmin=-0.5, xmax=1.5,
xtick style={color=black},
xtick={0,1},
xticklabels={Late,Early},
y grid style={darkgray176},
ylabel={AUC PR},
ymajorgrids,
ymin=0.6, ymax=0.8,
ytick style={color=black}
]

\draw[draw=none,fill=magenta,fill opacity=0.3] (axis cs:-0.2,0) rectangle (axis cs:0,0.7085);
\draw[draw=none,fill=magenta,fill opacity=0.3] (axis cs:0.8,0) rectangle (axis cs:1,0.725);
\addlegendimage{ybar,area legend,draw=none,fill=magenta,fill opacity=0.3} %postaction={pattern=north east lines}
\addlegendentry{No Concept}

%fill postaction={pattern=north east lines} or dots
\draw[draw=none,fill=magenta,fill opacity=0.7] (axis cs:-1.38777878078145e-17,0) rectangle (axis cs:0.2,0.736);
\draw[draw=none,fill=magenta,fill opacity=0.7] (axis cs:1,0) rectangle (axis cs:1.2,0.755);
\addlegendimage{ybar,area legend,draw=none,fill=magenta,fill opacity=1} %,postaction={pattern=dots}
\addlegendentry{Concept}

%\addlegendimage{area legend,pattern=north east lines,pattern color=magenta}
%\addlegendentry{Late}

%\addlegendimage{
%    area legend,
%    pattern=dots,
%    pattern color=magenta,
%    pattern/.style={dot size=6pt}
%}
%\addlegendentry{Early}

\path [draw=gray, semithick]
(axis cs:-0.1,0.69)
--(axis cs:-0.1,0.727);

\path [draw=gray, semithick]
(axis cs:0.9,0.705)
--(axis cs:0.9,0.745);

\path [draw=gray, semithick]
(axis cs:0.1,0.715)
--(axis cs:0.1,0.757);

\path [draw=gray, semithick]
(axis cs:1.1,0.733)
--(axis cs:1.1,0.777);

\addplot [semithick, gray, mark=-, mark size=5, mark options={solid}, only marks]
table[row sep=crcr]{%
-0.1 0.69\\
-0.1 0.727\\
0.9 0.705\\
0.9 0.745\\
0.1 0.715\\
0.1 0.757\\
1.1 0.733\\
1.1 0.777\\
};

\end{axis}

\end{tikzpicture}
}
    \hfill
   % This file was created with tikzplotlib v0.10.1.
\resizebox{0.48\columnwidth}{!}{
\begin{tikzpicture}

\definecolor{darkgray176}{RGB}{176,176,176}
\definecolor{gray}{RGB}{128,128,128}
\definecolor{lightgray204}{RGB}{204,204,204}
\definecolor{orange}{RGB}{255,165,0}

\begin{axis}[
legend cell align={left},
legend columns=3,
legend style={
  fill opacity=0.8,
  draw opacity=1,
  text opacity=1,
  at={(0.5,0.98)},
  anchor=north,
  draw=lightgray204
},
tick align=outside,
tick pos=left,
label style={font=\LARGE},
tick label style={font=\Large},
x grid style={darkgray176},
xmin=-0.5, xmax=1.5,
xtick style={color=black},
xtick={0,1},
xticklabels={Late,Early},
y grid style={darkgray176},
ylabel={AUC PR},
ymajorgrids,
ymin=0.2, ymax=0.6,
ytick style={color=black}
]

%fill \draw[draw=none,fill=orange,fill opacity=0.3,postaction={pattern=north east lines}] (axis cs:-0.2,0) rectangle (axis cs:0,0.375); \draw[draw=none,fill=orange,fill opacity=0.3,postaction={pattern=dots}] (axis cs:0.8,0) rectangle (axis cs:1,0.373);

\draw[draw=none,fill=orange,fill opacity=0.3] (axis cs:-0.2,0) rectangle (axis cs:0,0.375);
\draw[draw=none,fill=orange,fill opacity=0.3] (axis cs:0.8,0) rectangle (axis cs:1,0.373);
\addlegendimage{ybar,area legend,draw=none,fill=orange,fill opacity=0.3}
\addlegendentry{No Concept}

\draw[draw=none,fill=orange,fill opacity=0.7] (axis cs:-1.38777878078145e-17,0) rectangle (axis cs:0.2,0.4715);
\draw[draw=none,fill=orange,fill opacity=0.7] (axis cs:1,0) rectangle (axis cs:1.2,0.4845);
\addlegendimage{ybar,area legend,draw=none,fill=orange,fill opacity=1} %postaction={pattern=dots}
\addlegendentry{Concept}

%\addlegendimage{area legend,pattern=north east lines,pattern color=orange}
%\addlegendentry{Late}

%\addlegendimage{
%    area legend,
%    pattern=dots,
%    pattern color=orange,
%    pattern/.style={dot size=6pt}
%}
%\addlegendentry{Early}

\path [draw=gray, semithick]
(axis cs:-0.1,0.332)
--(axis cs:-0.1,0.418);

\path [draw=gray, semithick]
(axis cs:0.9,0.331)
--(axis cs:0.9,0.415);

\path [draw=gray, semithick]
(axis cs:0.1,0.419)
--(axis cs:0.1,0.524);

\path [draw=gray, semithick]
(axis cs:1.1,0.437)
--(axis cs:1.1,0.532);

\addplot [semithick, gray, mark=-, mark size=5, mark options={solid}, only marks]
table[row sep=crcr]{%
-0.1 0.332\\
-0.1 0.418\\
0.9 0.331\\
0.9 0.415\\
0.1 0.419\\
0.1 0.524\\
1.1 0.437\\
1.1 0.532\\
};

\end{axis}

\end{tikzpicture}
}
    \caption{Left: Video and audio as input modalities to early and late fusion models. Right: Text and audio as input modalities to early and late fusion models.}
    \label{fig:paired_modalities}
    \vspace{-0.5cm}
\end{figure}
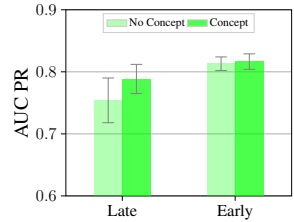
\begin{figure}[!ht]
    \centering
    \resizebox{0.5\linewidth}{!}{% This file was created with tikzplotlib v0.10.1.
\resizebox{0.9\columnwidth}{!}{
\begin{tikzpicture}

\definecolor{darkgray176}{RGB}{176,176,176}
\definecolor{gray}{RGB}{128,128,128}
\definecolor{lightgray204}{RGB}{204,204,204}
\definecolor{green}{RGB}{0,255,0}

\begin{axis}[
legend cell align={left},
legend columns=2,
legend style={
  fill opacity=1,
  draw opacity=1,
  text opacity=1,
  at={(0.5,0.98)},
  anchor=north,
  draw=lightgray204
},
tick align=outside,
tick pos=left,
label style={font=\LARGE},
tick label style={font=\Large},
x grid style={darkgray176},
xmin=-0.5, xmax=1.5,
xtick style={color=black},
xtick={0,1},
xticklabels={Late,Early},
y grid style={darkgray176},
ylabel={AUC PR},
ymajorgrids,
ymin=0.6, ymax=0.9,
ytick style={color=black}
]

\draw[draw=none,fill=green,fill opacity=0.3] (axis cs:-0.25,0) rectangle (axis cs:0,0.754);
\draw[draw=none,fill=green,fill opacity=0.3] (axis cs:0.75,0) rectangle (axis cs:1,0.813);
\addlegendimage{ybar,area legend,draw=none,fill=green,fill opacity=0.3} %,postaction={pattern=north east lines}
\addlegendentry{No Concept}

%postaction={pattern=north east lines} or dots postaction={pattern=dots}
\draw[draw=none,fill=green,fill opacity=0.7] (axis cs:0,0) rectangle (axis cs:0.25,0.7885);
\draw[draw=none,fill=green,fill opacity=0.7] (axis cs:1,0) rectangle (axis cs:1.25,0.8165);
\addlegendimage{ybar,area legend,draw=none,fill=green,fill opacity=0.7} %,postaction={pattern=dots}
\addlegendentry{Concept}

%\addlegendimage{area legend,pattern=north east lines,pattern color=green}
%\addlegendentry{Late}

%\addlegendimage{
%    area legend,
%    pattern=dots,
%    pattern color=green,
%    pattern/.style={dot size=6pt}
%}
%\addlegendentry{Early}
\path [draw=gray, semithick]
(axis cs:-0.125,0.718)
--(axis cs:-0.125,0.79);

\path [draw=gray, semithick]
(axis cs:0.875,0.802)
--(axis cs:0.875,0.824);

\path [draw=gray, semithick]
(axis cs:0.125,0.765)
--(axis cs:0.125,0.812);

\path [draw=gray, semithick]
(axis cs:1.125,0.804)
--(axis cs:1.125,0.829);

\addplot [semithick, gray, mark=-, mark size=5, mark options={solid}, only marks]
table[row sep=crcr]{%
-0.125 0.718\\
-0.125 0.79\\
0.875 0.802\\
0.875 0.824\\
0.125 0.765\\
0.125 0.812\\
1.125 0.804\\
1.125 0.829\\
};

\end{axis}

\end{tikzpicture}
}}
    \caption{Concept-based supervision for early and late fusion of all three video, audio, and text modalities.}
    \label{fig:three_modalities}
    \vspace{-0.5cm}
\end{figure}

\section{Conclusion}\label{sec:conclusion}

In this work, we address the challenge of improving model performance in multimodal video interpretation through concept-informed supervision. We demonstrated that the integration of concepts - which serve
as explaination for the final label - into the training
process improves the models performance while inherently creates a more interpretable framework. Leveraging MOByGaze, a recent video dataset with human-annotated multimodal concepts, we design an assessment strategy using Concept Modality Specific Datasets (CMSDs). 
These datasets are created by categorizing video samples based on specific modalities, such as visual, textual, or audio, that contain explanatory concepts for the given task. This categorization enables the evaluation of the discriminative power of multimodal models. Our experiments reveal that training with CMSDs significantly improves the performance of both late and early fusion models. Notably, this modality-informed supervision enables late fusion models to achieve comparable performance to early fusion models, highlighting the importance of detailed annotations in developing robust and self-explainable video interpretation model.

\bibliographystyle{IEEEbib}

{\small\bibliography{output.bbl}}

\end{document}